\pdfoutput=1

\documentclass[11pt]{article}

\usepackage[]{acl}

\usepackage{times}
\usepackage{latexsym}
\usepackage{booktabs}
\usepackage[T1]{fontenc}

\usepackage[utf8]{inputenc}

\usepackage{microtype}
\usepackage{amsmath}
\usepackage{hyperref}
\usepackage{graphicx}
\DeclareMathOperator*{\maxi}{max}
\title{Investigating the Impact of Pre-trained Language Models on Dialog Evaluation}  

\author{Chen Zhang$^{\dag,\star}$ \quad Luis Fernando D’Haro$^\ddag$ \quad Yiming Chen$^{\dag}$  \\ 
\textbf{Thomas Friedrichs}$^\star$ \quad \textbf{Haizhou Li}$^{\dag,\star\star}$     \\
  $^\dag$National University of Singapore \quad $^\star$Robert Bosch (SEA), Singapore \\
  $^\ddag$Universidad Politécnica de Madrid, Spain 
  \quad $^{\star\star}$Kriston AI Lab, China \\
        \tt \{chen\_zhang,yiming.chen\}@u.nus.edu, \\
  }

\begin{document}
\maketitle

\begin{abstract}
Recently, there is a surge of interest in applying pre-trained language models (Pr-LM) in automatic open-domain dialog evaluation. Pr-LMs offer a promising direction for addressing the multi-domain evaluation challenge. Yet, the impact of different Pr-LMs on the performance of automatic metrics is not well-understood. This paper examines 8 different Pr-LMs and studies their impact on three typical automatic dialog evaluation metrics across three different dialog evaluation benchmarks. Specifically, we analyze how the choice of Pr-LMs affects the performance of automatic metrics. Extensive correlation analyses on each of the metrics are performed to assess the effects of different Pr-LMs along various axes, including pre-training objectives, dialog evaluation criteria, model size, and cross-dataset robustness. This study serves as the first comprehensive assessment of the effects of different Pr-LMs on automatic dialog evaluation.
\end{abstract}

\section{Introduction}
\label{sec:intro}
Evaluation is crucial for monitoring the research progress of dialog systems~\citep{comp-assess}. Even though human evaluation is the most accurate way to assess the performance of a dialog system, the required expenses, and efforts restrict its application to large-scale dialog evaluation tasks. Therefore, automatic dialog evaluation (ADE) serves as an efficient alternative to human evaluation. An ideal ADE metric is expected to evaluate dialog systems of different domains efficiently and effectively. Realizing such a metric is a challenging task~\citep{comp-assess}. One promising trend is to leverage large-scale pre-trained language models (Pr-LMs)~\citep{bert,xlnet,roberta,electra}, which have gained significant momentum in a wide range of NLP tasks. Several recent studies~\citep{deep-amfm,usr,dscore,dynaeval} have also demonstrated their usefulness in ADE. 

However, different Pr-LMs have different pre-training schemes and they are not directly optimized for dialog evaluation. A dialog metric relies on dialog-specific features for determining the quality of dialog responses and whether the pre-training process of a Pr-LM captures such features has not been extensively studied. In addition, the choice of Pr-LM will significantly affect the ADE metrics' performance and generalizability across various evaluation tasks and evaluation dimensions (e.g., coherence, interestingness, naturalness). Currently, there is no comprehensive analysis to guide the choice of Pr-LM for ADE.

To this end, a systematic study on how different Pr-LMs affect the evaluation effectiveness of various ADE metrics is highly sought after. In this paper, we survey 8 existing state-of-the-art Pr-LM variants and analyze their impact on the performance of three typical ADE metrics, the embedding-based similarity measure, the normalized sentence-level log probability, and the context-response coherence metric. These three metrics are commonly used across multiple domains and can directly relate to three basic dimensions to evaluate a dialog system: adequacy (semantic similarity with existing references), naturalness, and context coherence. Through extensive correlation analysis on three dialog evaluation benchmarks, we try to link the properties associated with the Pr-LMs, such as model size, source of pre-training data, and learning objective, with the metrics' final performance. Our work serves as a first step into understanding the role of different Pr-LMs in automatic open-domain dialog evaluation. In addition, it will help guide future research in making more informed choices when applying pre-trained language models in various dialog evaluation tasks. Note that we are not proposing a new evaluation metric to advance the state-of-the-art, but rather studying the impact of different Pr-LMs on existing ADE metrics.

The paper is organized as follows: Section~\ref{sec:prlm} discusses the 8 different pre-trained language models. Section~\ref{sec:ade} talks about the three ADE metrics. Section~\ref{sec:ed} is the experiment results and corresponding analysis. Finally, Section~\ref{sec:conclusion} concludes the paper and lays out the future work.

\section{Pre-trained Language Models}
\label{sec:prlm}
In this section, we discuss eight state-of-the-art variants of Pr-LMs and group them based on their pre-training objectives. Specifically, there are five groups. The first four groups are token-level representation models: a) masked language modeling (Section.~\ref{subsec:mlm}), b) replaced token detection (Section.~\ref{subsec:rtd}), c) causal language modeling (Section.~\ref{subsec:clm}), and d) permutation language modeling (Section.~\ref{subsec:plm}). The final group consists of sentence-level representation models, which are learned with specific sentence-level objectives (Section.~\ref{subsec:slm}). All of the Pr-LMs are based on the transformer architecture~\citep{vaswani-et-al-2017}, which has greatly changed the NLP landscape in recent years. Contextualized embeddings derived from the Pr-LMs are beneficial to the evaluation metrics since they have been demonstrated to carry rich syntactic structure information~\citep{liu-et-al-syntax-2019} and semantic meanings~\citep{tenney-et-al-2019} of sentences. Recent studies have also shown that rich world knowledge is encoded in the parameters of the Pr-LMs~\citep{zhou-et-al-2020}. Such information helps the metric better determine the linguistic quality of the generated dialog responses.

\subsection{Masked Language Modeling (MLM)}
\label{subsec:mlm}
MLM is a self-supervised pre-training task proposed in~\citep{bert} whereby the language model randomly masks out some tokens from the input sequence and the learning objective is to predict the original vocabulary ids of the masked tokens. Formally, given a input text sequence of n tokens, $T = \{t_1,\ldots,t_n\}$, we corrupt $T$ into $\tilde{T}$ by randomly masking a portion of tokens. A language model parameterized by $\theta$ is trained to reconstruct $T$ by predicting the set of masked tokens $\tilde{t}$ conditioned on $\tilde{T}$:

\begin{equation}
    \maxi_{\mathbf{\theta}}\text{log}p_{\theta}(T|\tilde{T}) = \maxi_{\mathbf{\theta}}\sum_{i\in{C}}\text{log}p_{\theta}(\tilde{t_i}=t_i|\tilde{T}) 
\end{equation}
where C presents the set of indices of masked tokens. Typical examples in this category include BERT~\citep{bert}, RoBERTa~\citep{roberta}, DeBERTa~\citep{deberta} and ELECTRA~\citep{electra}. 

\subsubsection{BERT~\citep{bert}} 
BERT is currently the most fundamental Pr-LM and a must-have baseline in a wide range of NLP tasks. The backbone of BERT is a stack of transformer encoders, which is pre-trained with two learning objectives in a multi-task setting. The first objective is the aforementioned mask language modeling. The second objective is next sentence prediction (NSP). BERT is useful in dialog evaluation in the sense that it reconstructs the original input text leveraging bidirectional context. This leads to accurate estimation of the sentence-level probability rendering it useful for evaluating the fluency of the generated dialog responses. Several previous studies have questioned the necessity of the NSP objective~\citep{albert,roberta}. Hence, BERT's benefits to context coherence metric are questionable.

\subsubsection{RoBERTa~\citep{roberta}} 
RoBERTa is an optimized version of BERT. The backbone architecture and hyperparameters of RoBERTa are almost the same as BERT. Unlike BERT, RoBERTa is solely optimized with the MLM objective. It has been shown to demonstrate better performance across a wide range of natural language understanding tasks compared to BERT. Since RoBERTa is only pre-trained with the MLM objective, we hypothesize that it may perform worse when evaluating sentence-level dialog properties, such as context coherence. However, compared to BERT, it may provide a more accurate estimation of the sentence-level probability since RoBERTa is a highly optimized token-level language model with the MLM objective.


\subsubsection{DeBERTa~\citep{deberta}} 
DeBERTa is a relatively new member in the BERT family\footnote{Introduced in the Ninth International Conference on Learning Representations (ICLR 2021)} and it improves upon BERT and RoBERTa by disentangling the attention mechanism whereby two separate vectors are used to represent each input token and they encode the token's semantic meaning and position in the text sequence accordingly. Then, disentangled matrices are adopted for computing the attention weights among all the tokens in the input sequence on their contents and relative positions, respectively. Furthermore, to account for the syntactical nuances in the input text, DeBERTa incorporates absolute position token embeddings right before the prediction layer. The model predicts the masked tokens based on aggregated information of the token contents and positions. Since DeBERTa only modifies the model architecture and there is no change in the pre-training objective. Hence, we hypothesize that applying DeBERTa for dialog evaluation will obtain similar performance w.r.t to the use of RoBERTa or BERT. 


\subsection{Replaced Token Detection (RTD)}
\label{subsec:rtd}
The RTD objective is introduced to pre-train the \textbf{ELECTRA} framework~\citep{electra}, which consists of an MLM-based generator and a discriminator. The backbone of ELECTRA is BERT with a binary classification layer on top. The difference between RTD and MLM is that MLM is a fill-in-the-blank task while RTD relies on the mask-and-infill mechanism. Specifically, some tokens in the input text sequence are replaced with alternatives sampled from a small generator. Then, a discriminator is trained to predict the identity of each of the tokens in the input sequence (whether it is the original token or a sampled one). Compared to MLM, the RTD pre-training objective is more useful to dialog evaluation because: 1) there is no mismatch between pre-training and testing as the model doesn't have to deal with the artificial mask tokens. 2) RTD directly optimizes the discriminator to distinguish tokens from the original distribution against adversarial samples from a generator conditioned on the bidirectional context. A major goal of dialog evaluation is to rank dialog responses sampled from different generators of varying degrees of quality conditioned on the dialog context. Hence, RTD better aligns with the goal of dialog evaluation compared to MLM. We hypothesize that applying ELECTRA as the backbone will achieve good performance for ADE.   

\subsection{Causal Language Modeling (CLM)}
\label{subsec:clm}
CLM objective is the traditional unidirectional autoregressive way of pre-training language models whereby the model tries to predict the next token conditioned on all the previous tokens. Formally, given a input text sequence of n tokens, $T = \{t_1,\ldots,t_n\}$, the language model parameterized by $\theta$ performs pre-training by maximizing the likelihood: 

\begin{equation}
    \maxi_{\mathbf{\theta}}\text{log}p_{\theta}(T) = \sum_{i=1}^{n}\text{log}p_{\theta}(t_i|t_{<i}) 
\end{equation}
In our study, we examine one Pr-LM with the CLM objective, \textbf{DialoGPT}~\citep{dialogpt}. The backbone architecture of DialoGPT is the same as GPT-2~\citep{gpt2}, which consists of a stack of masked multi-head self-attention layers. Unlike GPT-2, which is pre-trained on a massive amount of web-text data, DialoGPT is pre-trained on large-scale dialogs extracted from Reddit. The input to the DialoGPT is the context-response pairs and the model is pre-trained to maximize the probability of the response conditioned on the corresponding context. Out of all the Pr-LMs examined in this paper, DialoGPT is the only dialog-specific Pr-LM. Given its pre-training objective, we hypothesize that DialoGPT is useful to perplexity-based evaluation metrics.

\subsection{Permutation Language Modeling (PLM)}
\label{subsec:plm}

In~\citep{xlnet}, the authors propose the \textbf{XLNET} framework and introduce the PLM pre-training objective, which tries to combine the best of both CLM and MLM pre-training schemes: MLM-based Pr-LMs can capture bidirectional contextual information whereas CLM-based Pr-LMs don't assume the independence of tokens within the sequence and hence, can model the high-order and long-range dependency in natural language. With the PLM objective, the model aims to maximize the expected log likelihood of a sequence w.r.t. all its possible permutations of the factorization order. Although PLM addresses the shortcomings of MLM and CLM,~\citep{roberta} has demonstrated that with the same amount of pre-training data, XLNET is not superior compared to BERT. Hence, we hypothesize that its contribution to dialog evaluation may not be better than Pr-LMs optimized with the MLM objective.    

\subsection{Sentence-level Representation Learning}
\label{subsec:slm}

Compared to token-level representation models, sentence-level representation models may be more pertinent to the dialog evaluation tasks for the following reasons: 1) a dialog is essentially a coherent structure consisting of multiple utterances. The dynamics of information exchange among the interlocutors is captured by examining the interaction among utterances instead of a flattened sequence of tokens, which is too fine-grained. 2) Both the embedding-based similarity measure and the context-response coherence measure operate at the sentence level. Extra adaptation may be required for Pr-LMs pre-trained with the token-level objectives.

\subsubsection{Sentence-BERT~\citep{sentence-bert}} 

The backbone framework of Sentence-BERT (SBERT) is the BERT model. A siamese network is constructed to encode pairs of sentences. The model is then fine-tuned with the combination of SNLI~\citep{snli} and MNLI~\citep{mnli} datasets. The standard cross-entropy classification objective is adopted to optimize the model. Two other supervised objective functions that operate at the sentence level have also been experimented with. One is the mean-squared loss function and the other is the triplet loss function. Since SBERT is a sentence-level representation model, we hypothesize its performance will be better than BERT or RoBERTa for the embedding-based similarity and the context-response coherence metrics.

\subsubsection{SimCSE~\citep{simcse}} 
SimCSE is the current state-of-the-art in both supervised and unsupervised sentence representation learning. The unsupervised SimCSE only leverages dropout for data augmentation whereby the same sentence is passed into the encoder twice. With an independently sampled dropout mask, a positive pair can be obtained.
Other sentences in the same batch serve as negative instances. A contrastive loss is applied to pull the positive pairs closer and the negative pairs apart in the vector space. The supervised SimCSE uses the same contrastive loss on the entailment and contradiction pairs from the NLI datasets for sentence representation learning. Since SimCSE demonstrates state-of-the-art performance in a wide array of semantic textual similarity and transfer tasks, we hypothesize that it will also greatly benefit the dialog evaluation tasks and obtain better performance as compared to SBERT.

\section{Automatic Dialog Evaluation Metrics}
\label{sec:ade}

In this section, we introduce three simple, but widely-adopted automatic dialog evaluation metrics of which the choice of Pr-LM can lead to a significant impact on performance\footnote{Implementation at~\url{https://github.com/e0397123/dstc10_metric_track}}. 

\subsubsection{Embedding-based Similarity}

The embedding-based similarity metric (ESM) is a reference-based measure, which evaluates a generated dialog response based on its similarity w.r.t a reference sentence written by the human annotators. Usually, cosine similarity between the response embedding, $\mathbf{h}$ and the reference embedding, $\mathbf{r}$ is used as the metric score:

\begin{equation}
    sim(\mathbf{h}, \mathbf{r}) = \frac{\mathbf{h}^T\mathbf{r}}{||\mathbf{h}||\cdot||\mathbf{r}||}
\end{equation}
Compared to lexical-overlap metrics, such as BLEU~\citep{bleu} and ROUGE~\citep{rouge}, ESM is more flexible by allowing variations in the lexical form and focusing on the sentence-level semantics. The choice of embeddings has a significant impact on the performance of ESM. In our study, we investigate which Pr-LM provides useful vector representations to ESM. Several prior studies~\citep{bertscore,bleurt,deep-amfm} have proposed improvement versions of ESM leveraging Pr-LMs, but they didn't conduct a comprehensive analysis of the effects of different Pr-LM variants.

\subsubsection{Normalized Sentence-level Log Probability} Language models estimate the true distribution of natural language and assign probabilities to sequences of words. The sentence-level log probability (SLP) estimated by a language model can indicate the naturalness of a generated dialog response. Unlike ESM, SLP is a type of reference-free metrics, which doesn't depend on human-written references to determine the quality of generated dialog responses. In dialog evaluation, we want to examine how the naturalness of the generated dialog responses is affected by the corresponding context. Hence, we formulate SLP by taking into account the preceding contexts, $p$ of the generated response, $h$. For language models pre-trained with the CLM or PLM objectives, the normalized log probability is estimated as follows:
\begin{equation}
    \frac{1}{M}\text{log}p_{\theta}(C) = \frac{1}{M}\sum_{i=1}^{M}\text{log}p_{\theta}(c_i|c_{<i}) 
\end{equation}
where $C$ denotes the concatenation of $p$ and $h$. M is the total number of tokens in $C$. For language models pre-trained with the MLM objective, the normalized log probability is computed with:
\begin{equation}
    \frac{1}{M}\text{log}p_{\theta}(C) = \frac{1}{M}\sum_{i=1}^{M}\text{log}p_{\theta}(c_i|\tilde{C})
\end{equation}
where $\tilde{C}$ is the corrupted $C$ with $c_i$ being masked.

\subsubsection{Context-response Coherence} Discourse coherence is a broad area of research and in dialog evaluation, we try to assess coherence at different granularity. One is the coherence of the entire dialog flow~\citep{dynaeval} and the other is the local coherence at the turn-level, i.e., context-response coherence~\citep{cervone-2020}.  In our study, we focus on the local coherence assessment and adopt a simple metric (CoSim) to evaluate the coherence between the context, $p$, and the response, $h$: 
\begin{equation}
    sim(\mathbf{p}, \mathbf{h}) = \frac{\mathbf{p}^T\mathbf{h}}{||\mathbf{p}||\cdot||\mathbf{h}||}
\end{equation}
Unlike existing state-of-the-art model-based metrics, of which the evaluation capability may be jointly influenced by several different factors, such as learning strategy, training data, and model architecture, CoSim's performance heavily relies on the choice of sentence embeddings, and hence, it helps us straightforwardly examine the effects of Pr-LMs without the need to decouple impact due to other factors.






\section{Experiment \& Analysis}
\label{sec:ed}

This section demonstrates our key findings and is organized as follows: Section~\ref{subsec:deb} briefly describes the three dialog evaluation benchmarks we use. In section~\ref{subsec:ia}, we conduct preliminary analysis on the eight Pr-LMs' performance along axes, including model size and cross-dataset robustness to select the top-ranked Pr-LMs for further analysis in the subsequent sections. Section~\ref{subsec:rpm} includes the main results of Pr-LMs based on average turn-level Spearman rank correlations over all three evaluation benchmarks for each automatic metric. Section~\ref{subsec:fae} zooms into the USR-TopicalChat benchmark and analyzes the performance of the top-ranked Pr-LMs along each dialog evaluation dimension for each automatic metric. \textbf{Note that all the Pr-LMs are not fine-tuned with any task-specific datasets in our experiments}.

\subsection{Dialog Evaluation Benchmarks}
\label{subsec:deb}
We conduct our experiments on the Dailydialog-Eval~\citep{dailydialogeval}, USR-PersonaChat~\citep{usr} and USR-TopicalChat~\citep{usr} dialog evaluation benchmarks. The detailed statistics\footnote{In our experiment, we use the original human response as the reference w.r.t the dialog context. Hence, the number of data points in each dataset is less than the original amount.} are presented in table~\ref{tab:data-stats}. We select these three benchmarks for the following reasons: 

(1) They can be used for both reference-based and reference-free evaluation due to the presence of human-written references. 

(2) Annotations of multiple dialog evaluation dimensions are available. This enables a more fine-grained analysis of how different Pr-LMs affect individual dimensions. In both USR-Topical and USR-PersonaChat, each context-response pair is annotated by three dialog researchers along six evaluation dimensions based on different Likert scales: understandability (0-1), naturalness (1-3), maintaining context (1-3), interestingness (1-3), using knowledge (0-1) and overall quality (1-5). For Dailydialog-Eval, 900 dialog context-response data points are annotated and each data point is annotated by 4 Amazon Mechanical Turkers. The turkers rate the response along four different dimensions on a 5-point Likert scale: content, grammar, relevance, and overall. 

(3) The three benchmarks cover the three most common dialog domains often used in open-domain dialog system training. 



\begin{table*}[!t]
\centering
\resizebox{0.9\linewidth}{!}{
\begin{tabular}{@{}c|ccccc@{}}
\toprule
Dataset Name & No. Data & Utts Per Data & Words Per Utt & No. Annotations & Domain \\ \midrule
Dailydialog-Eval  & 800 & 4.9 & 20.18 & 128,00 & Chit-chat\\
USR-TopicalChat  & 300 & 11.20  & 23.14 & 5,400 & Knowledge-based\\
USR-PersonaChat  & 240 & 4.72 & 12.39 & 4,320 & Persona-based\\ \bottomrule
\end{tabular}
}
\caption{The statistics of Dailydialog-Eval, USR-TopicalChat and USR-PersonaChat.}
\label{tab:data-stats}
\end{table*}

\subsection{Initial Analysis}
\label{subsec:ia}


We conduct initial analysis on the Pr-LMs and those with good performance across all the benchmarks are selected for more fine-grained analysis. 

\subsubsection{The Effects of Model Size} For most Pr-LMs, the model size doesn't have a significant influence on the performance of the metrics. The only exception is RoBERTa-large vs RoBERTa-base for SLP where RoBERTa-large outperforms RoBERTa-base by 3.96 percent in terms of the average turn-level Spearman correlation over all the three evaluation benchmarks. This may be related to the fact that RoBERTa is solely optimized with the MLM objective. With more pre-training data and a larger size, the Pr-LM will provide a more accurate estimation of the sentence-level probability. Hence, in our subsequent analysis, we focus on the large version of the Pr-LMs. The detailed results of all Pr-LM variants can be found at~\url{https://bit.ly/2UFjWOH}.


\begin{table*}[!t]
\resizebox{\linewidth}{!}{
\begin{tabular}{@{}c|cccccccc@{}}
\toprule
Dataset & XLNET & DialoGPT & RoBERTa & BERT & DeBERTa & ELECTRA & SimCSE & SBERT\\ \midrule
Dailydialog-Eval  & 4.21$^*$ & 11.70 & 8.06$^*$ & 13.66 & 15.34 & 9.47$^*$ & \textbf{18.24} & 13.60 \\
USR-TopicalChat  & 17.61 & 11.11 & 18.13 & 22.17 & 20.74 & \textbf{37.81} & 22.89 & 26.22 \\
USR-PersonaChat  & 8.47$^*$ & 10.43$^*$ & 8.15$^*$ & 14.87 & 14.95 & 14.20 & \textbf{20.61} & 15.03 \\
Average & 10.10$^*$ & 11.08 & 11.45 & 16.90 & 17.01 & 20.49 & \textbf{20.58} & 18.28 \\ \bottomrule
\end{tabular}
}
\caption{Unweighted average turn-level Spearman correlation scores (\%) of Pr-LMs across two evaluation metrics (ESM \& CoSim) as well as across evaluation dimensions on each benchmark. The best score for each benchmark is highlighted in bold. $^*$denotes statistically insignificance ($p$-value $> 0.05$)}
\label{tab:per-dataset-corr}
\end{table*}

\subsubsection{Cross-dataset Robustness} We further analyze the cross-dataset robustness of each Pr-LM by examining their results for each of the evaluation benchmarks. Table~\ref{tab:per-dataset-corr} shows the per-dataset turn-level average Spearman correlation scores for all the Pr-LMs\footnote{SLP correlations are not included in the computation as sentence-level Pr-LMs cannot serve as the backbone of SLP.}. \textbf{A large variation in terms of average Spearman correlations can be observed}. The difference between the best-performing Pr-LM (SimCSE) and the worst-performing Pr-LM (XLNET) is 10.48 \%. This may be because SimCSE is optimized for natural language understanding tasks and thus, provides good semantic representation of the sentences while XLNET or DialoGPT is optimized to generate more fluent texts. Our evaluation task benefits from better semantic representations of the dialogue utterances. 

Furthermore, it can be seen that \textbf{SimCSE performs the best on Dailydialog-Eval and USR-PersonaChat while ELECTRA performs the best on both USR-TopicalChat}. \textbf{SimCSE is the most robust Pr-LM} as its performance is consistently good across all three benchmarks. The consistent performance of SimCSE makes it a good choice for multi-domain dialog evaluation metrics.

In general, \textbf{almost all of the Pr-LMs perform the best on USR-TopicalChat}. This is because the pre-training data domain of the Pr-LMs is close to that of USR-TopicalChat. Most of the Pr-LMs are pre-trained with Wikipedia articles and USR-TopicalChat contains dialogs discussing topics and facts from Wikipedia.

On the contrary, XLNET and DialoGPT do not perform as well as Pr-LMs pre-trained with the masked language modeling objective, such as BERT and DeBERTa. This may be because the bidirectional language models provide a more accurate representation of sentence semantics compared to uni-directional language models. More accurate representation of sentence meanings will greatly benefit the embedding-based metrics.

\subsection{Rankings of Pre-trained Language Models}
\label{subsec:rpm}

\begin{table*}[!t]
\centering
\resizebox{0.8\linewidth}{!}{
    \begin{tabular}{c|cccc}
    \toprule
     Pr-LM & ESM (Adequacy) & SLP (Fluency) & CoSim (Coherence) & Average \\ 
    \midrule
    BERT & 17.80 & \textit{15.56} & 16.01 & 16.46 \\
    DeBERTa & \underline{18.35} & 8.32$^*$ & 15.67 & 14.11 \\
    DialoGPT & 13.13 & 10.55$^*$ & 9.02 & 10.90$^*$ \\
    ELECTRA & \textit{18.45} & 6.58$^*$ & \textbf{22.53} & 15.85 \\
    RoBERTa & 12.38 & \textbf{18.37} & 10.52$^*$ & 13.75 \\
    SBERT & 17.57 & - & \underline{19.00} & - \\
    SimCSE & \textbf{21.68} & - & \textit{19.48} & - \\
    XLNET & 8.57$^*$ & \underline{10.85$^*$} & 11.62$^*$ & 10.35$^*$ \\ \bottomrule
     
    \end{tabular}
    }
    \caption{Unweighted average turn-level correlation scores (\%) of Pr-LMs w.r.t ESM (adequacy metric), SLP (fluency metric) and CoSim (coherence metric) respectively. The best score for each metric is highlighted in bold. The second best is italicized and the third one is underlined. $^*$ denotes statistically insignificance ($p$-value $> 0.05$)}
    \label{tab:large-rankings}
\end{table*}

After the initial analysis in section~\ref{subsec:ia}, we try to assess the impact of different Pr-LMs on the performance of each ADE metric. Table~\ref{tab:large-rankings} shows the average Spearman correlation scores of different PrLM-metric combinations. Each entry in the table is computed by taking the unweighted average of the corresponding correlation scores of all the three dialogue evaluation benchmarks. Based on the experiment results, we can make the following observations:

\subsubsection{Sentence-level vs Token-level} The results validate our hypothesis in section~\ref{subsec:slm} that \textbf{sentence-level representation models generally outperform the token-level representation models for the adequacy and coherence metrics}. For ESM (adequacy), SimCSE is the best and for CoSim (coherence), SimCSE and SBERT are among the top-3 rank models.

\subsubsection{RTD vs MLM} For CoSim (coherence)  and ESM (adequacy) metrics, we can observe that ELECTRA is ranked the first and the second respectively. It outperforms BERT by a significant margin of around 6 percent for CoSim (coherence). This validates our hypothesis in section~\ref{subsec:rtd} that \textbf{RTD equips the model with better discrimination power in determining responses of varying degrees of quality compared to MLM}.

\subsubsection{Impact of MLM} Generally, Pr-LMs pre-trained with mask language modeling (MLM) objective outperform the causal or permutation language models across all three metrics. Based on the results, it can be seen that \textbf{MLM-based Pr-LMs provide a more useful semantic representation compared to CLM-based or PLM-based models}. This may be attributed to MLM-based Pr-LMs' ability to capture bi-directional contextual information. In addition, RoBERTa and BERT are ranked the first and the second for the SLP metric, this validates our hypothesis in section~\ref{subsec:mlm} that \textbf{a highly optimized MLM-based Pr-LM is capable of providing an accurate estimation of sentence naturalness for dialog evaluation}. Moreover, DeBERTa's performance is not better than that of BERT. This is consistent with our hypothesis that \textbf{modifications to model architecture instead of pre-training objectives may not bring performance improvement in dialog evaluation}. 

\subsubsection{Impact of CLM/PLM}
Based on the average correlation scores, XLNET is the lowest-ranked model for ESM and CoSim metrics. This corroborates our hypothesis in section~\ref{subsec:plm} that the contribution of \textbf{PLM-based Pr-LM to dialog evaluation is not better than Pr-LMs optimized with the MLM objective}. Furthermore, it is surprising that DialoGPT performs poorly in terms of these three metrics. A possible reason is that even though DialoGPT is a dialog-specific language model, it is pre-trained with large-scale Reddit data, which is more casual and colloquial in style while the dialogs in the three benchmarks are written by humans to fulfill specific purposes. Hence, the language used may be more formal and the quality of the text is better compared to that of Reddit conversations. \textbf{Future work should explore adaptation techniques of Pr-LMs to perform different dialog evaluation tasks}.

\subsubsection{Correlation Across Metrics} \textbf{It can be observed that the correlations scores in the ESM (adequacy) and CoSim (coherence) categories are generally higher than those in SLP (fluency)}. This may be due to the properties of the dialog responses whereby a coherent and adequate response is generally fluent while a fluent response may be off-topic or irrelevant to the context. ESM (adequacy) and CoSim (coherence) are designed to distinguish relevant responses from irrelevant ones. They can detect a fluent, yet off-topic response. However, it is hard for SLP (fluency), which is specifically designed to evaluate the naturalness of the generated responses, to distinguish the good from the bad in such scenarios.

\subsection{Fine-grained Analysis on Evaluation Dimension}
\label{subsec:fae}   

Section~\ref{subsec:rpm} provides a holistic comparison of the Pr-LMs for automatic dialog evaluation. In this section, we analyze the performance of SimCSE, RoBERTa, and ELECTRA at a more fine-grained level on the USR-TopicalChat Benchmark. Table~\ref{tab:usrtopical-corr} showcases the Spearman correlation results at turn level. It can be seen that \textbf{ESM-ELECTRA combination performs the best in 3 out of all 6 evaluation dimensions. The average correlation scores of ESM-ELECTRA are even approaching that of state-of-the-art USR metric}, which is a model-based metric specifically optimized for the dialog evaluation task. Remarkably, in the ESM (adequacy) category, ELECTRA outperforms the second-best model, SimCSE, by an absolute 13 percent for the overall category. \textbf{ELECTRA's superior performance is due to its RTD pre-training objective since all three models are pre-trained on similar datasets and based on similar model architecture. Future work can consider adapting RTD to the sentence-level}. 

In addition, all three Pr-LMs perform quite well along the using knowledge dimension along the ESM (adequacy) and CoSim (coherence) categories even though no specific adaptation is performed to incorporate the external knowledge sources associated with the dialogs. This corroborates with findings in prior studies mentioned in section~\ref{sec:prlm} that \textbf{world knowledge is implicitly encoded in the parameters of the Pr-LMs. Therefore, applying Pr-LMs for multi-domain automatic dialog evaluation is a viable direction}.


\begin{table*}[!t]
\resizebox{\linewidth}{!}{
    \begin{tabular}{@{}c|ccc|cc|ccc|c@{}}
    \hline\noalign{\smallskip}
    \multicolumn{1}{c}{USR-TopicalChat} & \multicolumn{3}{c}{ESM (Adequacy)} & \multicolumn{2}{c}{SLP (Fluency)} & \multicolumn{3}{c}{CoSim (Coherence)} & \multicolumn{1}{c}{SoTA}\\
    \noalign{\smallskip}\hline\noalign{\smallskip}
    Dimension & RoBERTa  & SimCSE & ELECTRA & 
    RoBERTa & ELECTRA & RoBERTa  & SimCSE & ELECTRA & USR \\
    \noalign{\smallskip}\hline\noalign{\smallskip}
     Understandability & 12.60 & 17.94 & \textbf{34.27} & 20.50 & \underline{7.09$^*$} & 10.65$^*$ & 13.78 & 25.36 & 32.68 \\
     
     Naturalness  & 12.97 & 13.26 & \textbf{34.52} & 22.19 & 5.27$^*$ & 11.98 & 14.91 & 31.17 & 32.54\\
     
     Relevance & 20.22 & 25.43 & 34.78 & 26.79 & 2.47$^*$ & 20.37 & \underline{36.81}  & 36.10 & \textbf{37.69}\\ 
     
    Interestingness  & \underline{29.79} & \underline{37.69} & \underline{47.52} & \underline{29.05} & 5.18$^*$ & \underline{21.73} & 21.99 & \underline{\textbf{49.54}} & 48.77 \\
    
    Using Knowledge  & 26.79 & 31.66  & 32.52 & 16.82 & 5.11$^*$ & 6.94$^*$ & 5.88$^*$ & 37.17 & \textbf{44.68} \\
    
    Overall  & 25.78  & 32.83 & \textbf{45.89} & 27.89 & 1.56$^*$ & 17.76 & 22.52  & 44.83 & 41.92\\
    \noalign{\smallskip}\hline\noalign{\smallskip}
    Average & 21.36  & 26.47 & 38.42 & 23.87 & 4.45$^*$ & 14.91 & 19.32 &  37.36 & \textbf{39.71}\\
     \hline\noalign{\smallskip}  
    \end{tabular}
}
\caption{Turn-level Spearman correlation scores (\%) of each metric on the USR-TopicalChat benchmark along individual dialog evaluation dimensions are reported. For each metric, the results of RoBERTa, SimCSE, and ELECTRA are presented. USR~\citep{usr} is the state-of-the-art reference-free metric for this benchmark. The highest correlation score along each dimension is highlighted in bold.  The highest correlation score along each Pr-LM is underlined. Since SimCSE is a sentence-level representation model, it cannot be used for SLP. $^*$ indicates statistically insignificance ($p$-value $> 0.05$).}
\label{tab:usrtopical-corr}
\end{table*}

Furthermore, the interestingness dimension assesses whether a dialog response is generic/dull or specific to the context and the relevance dimension determines whether a dialog response is on-topic or off-topic w.r.t the corresponding context. \textbf{Even though RoBERTa, SimCSE, and ELECTRA are not directly pre-trained to determine the interestingness and relevance of a dialog response, they perform well when used as the backbones of CoSim (coherence) and ESM (adequacy)}. CoSim-ELECTRA even outperforms USR (49.54 vs 48.77) along the interesting dimension. The reason may be that for ESM (adequacy), there is the presence of ground-truth references, which are not dull nor off-topic. CoSim (coherence) is explicitly designed to look into the context. ELECTRA, RoBERTa, and SimCSE are state-of-the-art semantic representation models. When the meaning of the sentences is accurately encoded, the performance of both metrics will be greatly boosted.

When evaluating responses with ESM (adequacy) and CoSim (coherence) along understandability and naturalness, RoBERTa and SimCSE don't perform as well as ELECTRA. Their performance is also worse than when applied to evaluate along other dimensions. However, \textbf{ESM-ELECTRA performs exceptionally well along these two dimensions and it even outperforms USR. This showcases that ELECTRA may be a good and robust candidate for future development of embedding-reliant ADE metrics}.

However, when used for SLP (fluency), ELECTRA's performance is far worse compared to when used for ESM (adequacy) and CoSim (coherence). The RTD pre-training objective of ELECTRA is to optimize the discriminator instead of the MLM-based generator. Using ELECTRA's generator as a language model to estimate the sentence-level log probability may not be as accurate as RoBERTa, which has been highly optimized for the MLM objective. This explains ELECTRA's poor performance for the SLP metric.


\section{Conclusion \& Future Work}
\label{sec:conclusion}

In conclusion, this paper provides a comprehensive assessment of the impact of 8 different state-of-the-art Pr-LMs on ADE metrics. We try to analyze how different pre-training objectives align with the dialog evaluation task. Through extensive correlation analysis, we find out that sentence-level representation models are more robust for multi-domain evaluation tasks. ELECTRA is good at distinguishing the relevant or specific responses from the off-topic or dull responses. Finally, MLM-based Pr-LMs work better than CLM/PLM-based Pr-LMs on evaluating the fluency aspect of the responses. In the future, we will adapt the token-level RTD objective to sentence-level to better align with the dialog evaluation task. Additionally, with the insights from this work, we will try to propose new metrics that are useful to multi-domain dialog evaluation. Lastly, we will further examine the impact of task-specific fine-tuning of different Pr-LMs on automatic dialog evaluation metrics. 

\bibliography{custom}
\bibliographystyle{output.bbl}

\end{document}